\documentclass{article}
\usepackage{ijcai19}

\usepackage{times}
\usepackage{soul}
\usepackage{url}
\usepackage[utf8]{inputenc}
\usepackage[small]{caption}
\usepackage{graphicx}
\usepackage{amsmath}
\usepackage{booktabs}
\usepackage{mdframed}
  \usepackage[]{algorithm2e}
  \usepackage{graphicx}
  \usepackage{subcaption}
\usepackage{balance}

\usepackage{booktabs}
\usepackage[table,xcdraw]{xcolor}
  \usepackage{natbib}
  \usepackage{har2nat}
  \usepackage{mathrsfs}
\newcommand{\possessivecite}[1]{\citeauthor{#1}'s \citeyear{#1}}

\title{IJCAI--19 Example on typesetting multiple authors}

 \author{
Selmer Bringsjord$^1$ \and
Naveen Sundar Govindarajulu$^2$
\affiliations
$^{1,2}$Rensselaer Polytechnic Institute (RPI)
\\ Troy NY 12180 USA
\emails
selmer.bringsjord@gmail.com,
naveensundarg@gmail.com,
}

\usepackage{mathrsfs}
\usepackage{pslatex}
\usepackage{booktabs}
\usepackage{verbatim}
\usepackage{color}
\usepackage{amsfonts}
\usepackage{amsmath}
\usepackage{amssymb}

\usepackage{xcolor}

\usepackage{proof}
\usepackage{fancybox}
\usepackage{bussproofs}

\mathchardef\mhyphen="2D

\newcommand{\lsort}[1]{%
  \ensuremath{\mbox{\textsf{#1}}}}
\newcommand{\defsort}[2]{%
  \newcommand{#1}{\lsort{#2}}}

\defsort{\Action}{Action}
\defsort{\Time}{Time}
\defsort{\Self}{Self}
\defsort{\SortName}{SortName}

\defsort{\Agent}{Agent}
\defsort{\Entrant}{Entrant}
\defsort{\ActionType}{ActionType}
\defsort{\Moment}{Moment}
\defsort{\Boolean}{Boolean}
\defsort{\PayOut}{PayOut}
\defsort{\Fluent}{Fluent}
\defsort{\Event}{Event}
\defsort{\Object}{Object}
\defsort{\Numeric}{Numeric}

\newcommand{\lsymbol}[1]{%
  \ensuremath{\mathit{#1}}}
\newcommand{\defsymbol}[2]{%
  \newcommand{#1}{\lsymbol{#2}}}
\defsymbol{\action}{action}
\defsymbol{\initially}{initially}
\defsymbol{\holds}{holds}
\defsymbol{\mirrored}{mirrored}
\defsymbol{\happens}{happens}
\defsymbol{\clipped}{clipped}
\defsymbol{\initiates}{initiates}
\defsymbol{\terminates}{terminates}
\defsymbol{\prior}{prior}
\defsymbol{\interval}{interval}
\defsymbol{\refrain}{refrain}
\defsymbol{\harm}{harm}

\defsymbol{\does}{does}
\defsymbol{\plans}{plans}
\defsymbol{\act}{act}
\defsymbol{\react}{react}
\defsymbol{\payTot}{pay_{tot}}
\defsymbol{\fight}{fight}
\defsymbol{\coop}{coop}
\defsymbol{\enter}{enter}
\defsymbol{\stayout}{stayout}
\defsymbol{\learns}{learns}
\defsymbol{\redsplotched}{red\mhyphen splotched}
\defsymbol{\hassplotch}{has\mhyphen splotch}
\defsymbol{\payoff}{payoff}

\defsymbol{\capable}{capable}
\defsymbol{\destroyed}{destroyed}

\defsymbol{\removesplotch}{remove\mhyphen splotch}
\defsymbol{\remove}{remove}

\defsymbol{\cogito}{cogito}
\defsymbol{\named}{named}
\defsymbol{\deter}{deter}
\defsymbol{\enhance}{enhance}
\defsymbol{\attack}{attack}
\defsymbol{\cost}{cost}

\newcommand{\lconstant}[1]{%
  \ensuremath{\mbox{\textsf{#1}}}}
\newcommand{\defconstant}[2]{%
  \newcommand{#1}{\lconstant{#2}}}
\defconstant{\Enter}{Enter}
\defconstant{\StayOut}{StayOut}
\defconstant{\Fight}{Fight}
\defconstant{\Acquiesce}{Acquiesce}
\defconstant{\cs}{cs }
\defconstant{\estimate}{estimate}

\newcommand{\lmodality}[1]{%
  \ensuremath{\mathbf{#1}}}
\newcommand{\defmodality}[2]{%
  \newcommand{#1}{\lmodality{#2}}}
\defmodality{\common}{C}
\defmodality{\knows}{K}
\defmodality{\believes}{B}
\defmodality{\perceives}{P}
\defmodality{\mental}{M}
\defmodality{\ought}{O}

\defmodality{\desires}{D}
\defmodality{\intends}{I}
\defmodality{\says}{S}

\newcommand{\DCEC}{\ensuremath{\mathcal{DCEC}}}

\newcommand{\lif}{\Rightarrow}
\newcommand{\liff}{\Leftrightarrow}
\newcommand{\sep}{\ \lvert \ }

\defsymbol{\wipe}{wipe\mhyphen fore\mhyphen head}

\defsymbol{\us}{\mathsf{us}}
\defsymbol{\iran}{\mathsf{iran}}
\defsymbol{\israel}{\mathsf{israel}}
\defsymbol{\russia}{\mathsf{russia}}
\defsymbol{\T}{\mathsf{T}}
\defsymbol{\ugv}{\mathsf{ugv}}
\defsymbol{\uav}{\mathsf{uav}}
\defsymbol{\carrying}{carrying}
\defsymbol{\firefight}{firefight}
\defconstant{\main}{main}
\defsymbol{\mission}{mission}
\defconstant{\silence}{silence}
\defconstant{\now}{now}

\defconstant{\solda}{\ensuremath{\mathsf{soldier_A}}}
\defconstant{\soldb}{\ensuremath{\mathsf{soldier_B}}}
\defconstant{\sold}{\ensuremath{\mathsf{soldier}}}
\defconstant{\commander}{\ensuremath{\mathsf{commander}}}

\defconstant{\enemyterritory}{enemyterritory}
\defconstant{\baseb}{\ensuremath{\mathsf{base_b}}}
\defsymbol{\move}{move}
\defsymbol{\allowed}{\mathbf{allowed}}
\defconstant{\past}{past}
\defconstant{\jack}{jack}

\defsymbol{\rich}{rich}

\let\originalleft\left
\let\originalright\right
\renewcommand{\left}{\!\mathopen{}\mathclose\bgroup\originalleft}
\renewcommand{\right}{\aftergroup\egroup\!\originalright}

\defconstant{\robot}{Robot}

\newcounter{parens}
\def\countlparen{%
    \addtocounter{parens}{1} \overset{\ensuremath{{\color[rgb]{0.7,.7,0.7} ^{\the\value{parens}}}
}}{\lparen}%
}
\def\countrparen{%
\overset{\ensuremath{{\color[rgb]{0.7,.7,0.7} ^{\the\value{parens}}}
}}{\rparen}\addtocounter{parens}{-1}%
}
\let\lparen(
\let\rparen)
\begingroup
    \catcode`(\active
    \catcode`)\active
    \gdef\countparens{%
        \let(\countlparen
        \let)\countrparen
    }
\endgroup   
\newenvironment{nested parentheses}
{%
    \catcode`(\active
    \catcode`)\active
    \countparens
    \setcounter{parens}{0}%
}

\newcommand{\type}[1]{\textsf{#1}}

\begin{document}

\title{Learning \textit{Ex Nihilo}}
\date{\texttt{version 0224192000NY}}
\maketitle

\begin{abstract}
\noindent
This paper introduces, philosophically and to a degree formally, the
novel concept of learning \textit{ex nihilo}, intended (obviously) to
be analogous to the concept of creation \textit{ex nihilo}.  Learning
\textit{ex nihilo} is an agent's learning ``from nothing,'' by the
suitable employment of schemata for deductive and inductive reasoning.
This reasoning must be in machine-verifiable accord with a formal
proof/argument theory in a \textit{cognitive calculus} (i.e., roughly,
an intensional higher-order multi-operator quantified logic), and this
reasoning is applied to percepts received by the agent, in the context
of both some prior knowledge, and some prior and current interests.
Learning \textit{ex nihilo} is a challenge to contemporary forms of
ML, indeed a severe one, but the challenge is offered in the spirt of
seeking to stimulate attempts, on the part of non-logicist ML
researchers and engineers, to collaborate with those in possession of
learning-\textit{ex nihilo} frameworks, and eventually attempts to
integrate directly with such frameworks at the implementation level.
Such integration will require, among other things, the symbiotic
interoperation of state-of-the-art automated reasoners and
high-expressivity planners, with statistical/connectionist ML
technology.
\end{abstract}





\section{Introduction}
\label{sect:intro}

This paper introduces, philosophically and to a degree
logico-mathematically, the novel concept of learning \textit{ex
nihilo}, intended (obviously) to be analogous to the concept of
creation \textit{ex nihilo}.\footnote{No such assumption as that
creation \textit{ex nihilo} is real or even formally respectable is
made or needed in the present paper.} Learning \textit{ex nihilo} is
an agent's learning ``from nothing,'' by the suitable employment of
schemata for deductive and inductive reasoning.  This reasoning must
be in machine-verifiable accord with a formal proof/argument theory in
a \textit{cognitive calculus}, and this reasoning is applied to
percepts received by the agent, in the context of both some prior
knowledge, and some prior and current interests.  Roughly,
cognitive calculi include inferential components of intensional
higher-order multi-operator quantified logics, in which expressivity
far outstrips off-the-shelf modal logics and possible-worlds
semantics, and a number of such calculi have been introduced as bases
for AI that is unrelated to learning; e.g.\ see
\cite*{nsg_sb_dde_ijcai}.  The very first cognitive calculus, replete
with a corresponding implementation in ML, was introduced in
\cite*{ka_sb_scc_seqcalc}.)

Learning \textit{ex nihilo} is a challenge to contemporary forms of
ML, indeed a severe one, but the challenge is offered in the spirt of
seeking to stimulate attempts, on the part of non-logicist ML
researchers and engineers, to collaborate with those in possession of
learning \textit{ex nihilo} frameworks, and eventually attempts to
integrate directly with such frameworks at the implementation level.
Such integration will require, among other things, the symbiotic use
of state-of-the-art automated reasoners (such as ShadowReasoner, the
particular reasoner that for us powers learning \textit{ex nihilo})
with statistical/connectionist ML technology.




\section{A Starting Parable}
\label{sect:starting_parable}
  
Consider, for instance, Robert, a person of the human
variety\footnote{The concept of personhood is a \emph{mental} one that
rides well above such merely biological categories as \textit{Homo
sapiens sapiens}.  In a dash of insight and
eloquence, \citeasnoun{charniak.intro} declare that ``the ultimate
goal of AI is to build a person'' (p.\ 7) --- from which we can deduce
that personhood is in no way inseparably bound up with the particular
carbon-based human case.  The logico-computational modeling of
reasoning at the level of persons, crucial for learning \textit{ex
nihilo}, along with a synoptic account of personhood itself, is
provided in \cite{rascals_in_sun}.} who has just arrived for a
black-tie dinner party at a massive and manicured stone mansion to
which he has never been, hosted by a couple (who have told him they
own the home) he has never met, and is soon seated at an elegant
table, every seat of which is occupied by a diner Robert is now
meeting for the very first time.\footnote{Robert does know himself
(and in fact self-knowledge is essential for learning \textit{ex
nihilo}), but, again, he doesn't know any of the other diners.}  A
thin, tall, crystal glass of his (arrayed among three others, each of
a different shape, that are his as well) is gradually filled with
liquid poured from a bottle that he notices carries the words `Louis
Roederer,' which have no particular meaning for him; as the pour
unfolds, Robert notices tiny bubbles in the liquid in his glass, and
the white-tuxedoed server says, ``Your apertif, sir.''  At this point,
Robert is in position to learn an infinite number of propositions
\textit{ex nihilo}.  He has received no large dataset, and the only
direct communication with him has been composed solely of rather empty
pleasantries and the one perfunctory declaration that he has been
served an apertif.  Yet as Robert takes his first (stunning) sip of
what he has already learned \textit{ex nihilo} is expensive
Champagne,\footnote{Robert perceives that his beverage is sparkling
  wine, that it's likely quite dear, and knows enough about both the
  main countries that produce such a thing (viz.\ USA, Spain, France,
  and Italy), and linguistics, to reason to the belief that his
  beverage's origin is French, and hence that it's Champagne.} and as
he glances at the other five guests seated at the table, he is poised
to learn \textit{ex nihilo} without bound.  How much new knowledge he
acquires is simply a function of how much time and energy he is
willing and able to devote to the form of learning in question.  As
his water glass is filled, first with a wafer-thin slice of lemon
dropped in deftly with silver tongs, and then with the water itself,
he gets started:

For example, Robert now knows that his hosts find acceptable his
belief that they are quite wealthy.  [They may not in fact \emph{be}
  wealthy (for any number of reasons), but they know that Robert's
  perceiving what they have enabled him to perceive will lead to a
  belief on his part that they are wealthy, and Robert knows that they
  know this.]  Robert now also knows that the menu, on the wine side,
includes at least two additional options, since otherwise his array of
empty glasses wouldn't number three, one of which he knows is for
water.  $ldots$

\section{Learning \textit{Ex Nihilo} is Ubiquitous}
\label{sect:ubiquity}

Of course, where there is one parable, countless others can be found:
Herman isn't the black-tie kind of person.  Given a choice between the
dinner Robert had versus one under the stars in the wilderness,
prepared on an open fire, Herman will take the latter, every time.
Having just finished such a meal, Herman is now supine beside the
fire, alone, many miles from civilization in the Adirondack Park, on a
very chilly but crystal-clear summer evening.  Looking up at the
heavens, he gets to thinking --- and specifically gets to learning
(\textit{ex nihilo}, of course).  Herman knows next to nothing about
astronomy.  As a matter of fact, in general, Herman doesn't go in much
for academics, period.  He sees a light traveling smoothly, steadily,
and quickly across his field of vision.  Herman asks himself: What is
this thing?  He hears no associated sound.  He isn't inclined to take
seriously that this is an alien spacecraft --- unless what he is
seeing is a total anomaly.  Is it? he asks himself.  He waits and
looks.  There is another.  This seems routine, but if so, and if this
is a UFO, the papers would routinely be filled with UFO ``sightings,''
and so on; but they aren't.  So, no, not a UFO.  The light, he next
notes, is traveling too quickly to be a jet at high altitude, and in
the dark like this, no light pollution whatsoever, jets at high
altitude are hard to see.  Herman notes that the object, as it moves,
blocks out his view of stars behind it.  Ah!  Herman now knows that he
has just seen a satellite in orbit, and with that done once, before
the night is out he will see two more.  Herman never knew that you
could just lay down under these conditions and see satellites; he also
never knew that there are a lot of satellites up there, going around
Earth, but he reasons that since his experience is from one particular
spot on the surface of Earth, it is likely to be representative of any
number of other locations, and hence there must be many of these
satellites in orbit.  Herman has now come to learn many things, and
the night is still young.

Robert and Herman are at the tip of an infinite iceberg of cases in
which agents learn \textit{ex nihilo}, both in rather mundane fashion
at fancy dinners and campfire dinners, and in the more exotic cases
seen in fiction (witness e.g.\ the eerie ability of Sherlock Holmes to
quickly learn \textit{ex nihilo} myriad things when meeting people for
the first time, a recurring and prominent phenomena in
PBS' \href{https://www.pbs.org/wgbh/masterpiece/shows/sherlock/}{Sherlock}.).
Moreover, it turns out that learning \textit{ex nihilo} is not only
ubiquitous, but is also --- upon empirical investigation --- a very
powerful way to learn in the academic sphere, where forcing oneself to
be interested enough to ask oneself questions, and then attempt to
reason to their answers, can pay demonstrable academic
dividends \cite{chi.eliciting.self-explanation,vanlehn_self-explanation}.

\section{Learning \textit{Ex Nihilo} Produces Knowledge}
\label{sect:len_produces_knowledge}

Please note that while it may seem to the reader that
learning \textit{ex nihilo} is rather relaxed, free-wheeling, and
epistemically risky, the fact is that we have \emph{very} high
standards for declaring some process, whether implemented in a person
or a machine, to be learning.  Put with brutal simplicity here,
genuine learning of $\phi$ by an agent, for us, must result in the
acquisition of knowledge by the agent, and knowledge in turn consists
in the holding of three conditions, to wit: (1) the agent must believe
that $\phi$ holds; (2) must have cogent, expressible, surveyable
justification for this belief; and (3) $\phi$ must in fact hold.  This
trio constitute the doctrine that knowledge consists of justified true
belief; we shall abbreviate this doctrine as `\texttt{k=jtb}.'  By
\texttt{k=jtb}, which reaches back at least to Plato, most of what is
called ``learning'' in AI today (e.g.\ so-called ``deep learning'') is
nothing of the sort.\footnote{For an argument, with which we are
somewhat sympathetic, that contemporary ``machine learning'' fails to
produce knowledge for the agent that machine-``learns,'' see
(Bringsjord et al.\ \citeyear*{do_machine-learning_machines_learn}).}
But in the case of our Robert and Herman, conditions (1)--(3) obtain
with respect to all the new knowledge we have ascribed to them, and
this would clearly continue to be true even if we added \textit{ad
infinitum} propositions that they can come to learn \textit{ex
nihilo}, stationary physically, but moving mentally.

\section{Learning \textit{Ex Nihilo} Includes a Novel Solution to the 
         Vexing Gettier Problem}
\label{sect:gettier}


Since Plato it was firmly held by nearly all those who thought about
the nature of human knowledge that \texttt{k=jtb} --- until the
sudden, seismic publication of
\cite{gettier_problem}, which appeared to feature clear examples in
which \texttt{jtb} holds, but not \texttt{k}.  It would be quite fair
to say that since the advent of Gettier's piece, to this very day,
defenders of \texttt{k=jtb} have been rather stymied; indeed, it
wouldn't be unfair to say that not only such defenders, but in fact
all formally inclined epistemologists, have since the advent of
Gettier-style counter-examples been scurrying and scrambling about,
trying to pick up the pieces and somehow build up again a sturdy
edifice.  Our account of learning \textit{ex nihilo} includes a
formal-and-computational solution to the Gettier problem, which in
turn allows AIs built with our automated-reasoning technology
(described below) to acquire knowledge in accord with \texttt{k=jtb}.
But first, what is the Gettier problem?

\citeasnoun{gettier_problem} presents a scenario in which Smith has
``strong evidence'' for the proposition

\begin{itemize}
  \item[$f$] Jones owns a Ford.
\end{itemize}

\noindent The evidence in question, Gettier informs us, includes that
``Jones has at all times in the past within Smith's memory owned a
car, and always a Ford, and that Jones has just offered Smith a ride
while driving a Ford.''  In addition, Smith has another friend, Brown,
whose whereabouts are utterly unknown to Smith.  Smith randomly picks
three toponyms
and ``constructs the following three propositions.''

\begin{itemize}
  \item[$g$] Either Jones owns a Ford, or Brown is in Boston.
  \item[$h$] Either Jones owns a Ford, or Brown is in Barcelona.
  \item[$i$] Either Jones owns a Ford, or Brown is in Brest-Litovsk.
\end{itemize}

Of course, $\{f\} \vdash g$, $\{f\} \vdash h$, $\{f\} \vdash
i$.\footnote{We are here by the single turnstyle of course denoting
some standard provability relation in a standard, elementary
extensional collection of inference schemata, such as that seen in
first-order logic = $\mathscr{L}_1$, a logical system discussed below.
The disjunction is of course inclusive.}  ``Imagine,'' Gettier tells
us, ``that Smith realized the entailment of each of these propositions
he has constructed by'' $f$, and on that basis is ``completely
justified in believing each of these three propositions.''  Two
further facts in the scenario yield the apparent counter-example, to
wit: Jones doesn't own a Ford, and is currently driving a rented car;
and, in a complete coincidence, Brown is in fact in Barcelona.
Gettier claims, and certainly appears to be entirely correct in doing
so, that Jones doesn't know $h$, yet $h$ is true, Smith believes $h$,
and Smith is justified in believing $h$ --- which is to say
that \texttt{jtb} appears to be clearly instantiated!


Learning \textit{ex nihilo} includes an escape from Gettier:
Encapsulated to a brutal degree here, we gladly allow that the
characters like Smith in
\possessivecite{gettier_problem} cases \emph{do} have knowledge on the
basis of a \texttt{k=jtb}-\emph{style} account, but the knowledge in
question can be cast at any number of five levels, 1 (more likely than
not) the weakest and 5 (certain) the strongest.  Specifically, we hold
that Smith knows at a level of 1, because belief in these cases is
itself at a strength level of 1, and that's because the argument
serving as justification for belief in these cases only supports
belief at that level.  To our knowledge, this proposed solution to the
counter-examples in question is new, though there are echoes of it in
\cite{theory.of.knowledge2.chisholm}.\footnote{Echoes only.
  Chisholm's main moves are flatly inconsistent with ours.  E.g., his
  definition of the longstanding \texttt{jtb}-based concept of
  knowledge includes not merely that the agent is \emph{justified} in
  believing $\phi$, but the stipulation that $\phi$ is \emph{evident}
  for the agent \cite[p.\ 102]{theory.of.knowledge2.chisholm}.  And
  his modifications of the \texttt{j} condition in the \texttt{jtb}
  triad are internalist, whereas ours are externalist, inhering as
  they do in formal structures and methods.  Somewhat amazingly, the
  learning-\textit{ex nihilo} diagnosis and resolution of Gettier
  cases is assumed in the literature to be non-existent.  E.g., here
  is what's said about Gettier cases in what is supposed to be the
  non-controversial and comprehensive SEP: \begin{quote}
  Epistemologists who think that the JTB approach is basically on the
  right track must choose between two different strategies for solving
  the Gettier problem.  The first is to strengthen the justification
  condition to rule out Gettier cases as cases of justified belief.
  This was attempted by Roderick Chisholm;$^{12}$ $\dots$ The other is
  to amend the JTB analysis with a suitable fourth condition, a
  condition that succeeds in preventing justified true belief from
  being ``gettiered.''  Thus amended, the JTB analysis becomes a JTB+X
  account of knowledge, where the `X' stands for the needed fourth
  condition.  \cite[\S 3, ``The Gettier
  Problem'']{analysis_knowledge_sep} \end{quote} Yet the
  learning \textit{ex nihilo}-solution, while retaining
  the \texttt{jtb} kernel, is based on neither of these two different
  strategies.}  An AI-ready inductive logic that allows Gettier to be
  surmounted in this fashion is presented
  in \cite{uncertaintyized_cognitive_calculus}.

\section{On Logico-Mathematics of Learning \textit{ex Nihilo}}
\label{sect:logico-mathematics}

Is there a logico-mathematics of learning \textit{ex nihilo}?  If so,
what is it, at least in broad strokes?  The answer to the first of
these questions is a resounding affirmative --- but in the present
paper, intended to serve as an introduction to a new form of human
learning driven by reasoning, and concomitantly as a challenge to
learning-focused AI researchers (incl.\ and perhaps esp.\ those in AI
who pursue machine learning in the absence of reasoning carried out in
confirmable conformity with inference schemata), the reader is
accordingly asked to be be willing to rest content (at least
provisionally) with but an encapsulation of the logico-mathematics in
question, and references (beyond those in the previous \S) to prior
work upon which the formal theory of learning \textit{ex nihilo} is
based.  This should be appropriate, given that via the present paper
we seek to place before the community a chiefly philosophical
introduction to learning \textit{ex nihilo}.  We present the core of
the relevant logico-mathematics, starting with the next paragraph.
Our presentation presumes at least some familiarity with formal
computational logic (both extensional and
intensional\footnote{Roughly, extensional logic invariably assigns a
semantic value to formulae in a purely compositional way, and is ideal
for formalizing mathematics itself; this is why the logical systems
traditionally used in mathematical logic are such things as
first-order and second-order logic.  Such logical systems, in their
elementary forms, are of course covered in the main AI textbooks of
today, e.g.\ \cite{aima.third.ed,luger_ai_book_6thed}.  In stark
contrast, the meaning of a formula $\phi$ in an intensional logic
can't be computed or otherwise obtained from it and what it's composed
of.  For a simple example, if $\phi$ is $\psi \rightarrow \delta$, and
we're in (extensional) zeroth-order logic (in which $\rightarrow$ is
the material conditional), and we know that $\psi$ if false, then we
know immediately that $\phi$ is true.  But if $\phi$ is instead
$\mathbf{B}_a \psi$, where $\textbf{B}$ is an agent-indexed belief
operator in epistemic logic, and $\psi$ is what agent $a$ believes,
the falsity of $\psi$ doesn't at all guarantee any truth-value for
$\mathbf{B}_a \psi$.  Here, the belief operator is an intensional
operator, and is likely to specifically be a modal operator in some
modal logic.}) and late 20th- and 21st-century automated
deduction/theorem proving.  (Learning \textit{ex nihilo} is, as we
shall soon see, explicitly based upon automated reasoning that is
non-deductive as well, but automated non-deductive reasoning is
something we can't expect readers to be familiar with.)  For readers
in the field of AI who are strong in statistical/connectionist ML,
and/or reinforcement learning and Bayesian approaches/reasoning, but
weak in formal computational logic, in either or both of its deductive
and inductive forms, we recommend
\cite{inconsistency_godel_modal_ijcai2016,nsg_sb_dde_ijcai}, and then
working backwards through readily available textbooks and papers cited
in this earlier IJCAI-venue work, and in the next two sub-sections.

\subsection{Logical Systems and Learning \textit{Ex Nihilo}}

The concept of a \textit{logical system}, prominent in the major
result known as Lindstr\"{o}m's Theorem,\footnote{Elegantly covered
in \cite{ebb.flum.thomas.2nded}.} provides a detailed and rigorous way
to treat logics abstractly and efficiently, so that e.g.\ we can
examine and (at least sometimes) determine the key attributes that
these logics have, relative to their expressive power.  A logical
system $\mathscr{L}$ is a triple
$$\langle \mathcal{L}, \mathcal{I}, \mathcal{S} \rangle$$
whose elements are, in turn, a formally specified language
$\mathcal{L}$ (which would customarily be organized in ways that are
familiar in programming languages; e.g.\ types would be specified); an
inference theory $\mathcal{I}$ (which would be a proof theory in the
deductive case, an argument theory in the inductive case, and best
called an \textit{inference theory} when inference schemata from both
categories are mixed) that allows for precise and machine-checkable
proofs/arguments, composed of inference schemata; and some sort of
semantics $\mathcal{S}$ by which the meaning of formulae in
$\mathcal{L}$ are to be interpreted.

Each of the elements of the abstract triple the individuates a given
logical system can be vast and highly nuanced, and perhaps even a
substantive part of a branch of formal logic in its own right.  For
example, where the logical system is standard first-order logic
$\mathscr{L}_1$, $\mathcal{S}$ will include all of established model
theory for first-order logic.  Lindstr\"{o}m's Theorem tells us
(roughly) that any movement to an extensional logical system whose
expressive power is beyond $\mathscr{L}_1$ will lack certain
meta-attributes that many consider desirable.  For instance,
second-order logic $\mathscr{L}_2$ isn't complete, whereas
$\mathscr{L}_1$ is.  This is no way stops AI researchers from working
on and with higher-order extensional logics.\footnote{The formal
verification of G\"{o}del's famous ontological argument for God's
existence, an argument that employs $\mathscr{L}_3$, has been verified
by AI researchers; see e.g.\ \cite{ecai2014_paleo_godel_proof}.}

For present learning \textit{ex nihilo}, the most important element in
the triple that makes for a logical system is $\mathcal{I}$, which can
be viewed as a set of inference schemata.\footnote{For simple logical
systems, the phrase `inference rules' is often used instead of the
more accurate `inference schemata,' and in fact there is a tradition
in places of using the former.  Because even an individual inference
schema can be quite complex, and can involve meta-logical constructs
and computation, talk of schemata is more accurate.  For instance, a
common inference schema in so-called natural deduction is
\begin{center}
\begin{tabular}{c}
$\phi(a),$ where $a$ is a constant in formula $\phi$\\
\hline
$\exists x \phi(x)$
\end{tabular}
\end{center}
but all sorts of further restrictions can be (and sometimes are)
placed on $\phi(a)$, such as that it must be a $\Delta_0$ formula.  As
such things grow more elaborate, it quickly makes precious little
sense to call them ``rules,'' and besides which many think of them as
programs.}  The reason is that learning \textit{ex nihilo} is based on
reasoning in which each inference is sanctioned by some
$I \in \mathcal{I}$, and on the \emph{automation} of this reasoning,
including when the inference schemata are non-deductive.  In computer
science and AI, a considerable number of people are familiar with
automated \emph{deductive} reasoners; after all, Prolog is based on
automated deductive reasoning, using only one inference schema
(resolution), involving formulae in a fragment of $\mathscr{L}_1$.
Learning \textit{ex nihilo}, in contrast, is based on automated
reasoning over \emph{any} inference schemata --- not only deductive
ones, but inductive ones, e.g.\ ones that regiment analogical
reasoning, abductive reasoning, enumerative inductive reasoning, and
so on.  All the reasoning patterns seen in inductive logic, in their
formal expressions, are possible as inference schemata employed in
learning \textit{ex nihilo}.\footnote{For a particular example of
formal, automated reasoning that blends deduction with analogical
reasoning, see \cite{GIfromLP_ijcai13}.  For a readable overview of
inference patterns in inductive logic that we formalize and automate,
see \cite{intro_inductive_logic_johnson}.}

\subsection{From Logical Systems to Cognitive Calculi}
\label{subsect:automated_reasoning_planning}

Because learning \textit{ex nihilo} frequently involves the mental
states of other agents (as seen e.g.\ in the parable regarding
Robert), we employ a novel class of logical systems called
\textit{cognitive calculi}, and they form part of the singular basis
of this new kind of learning.  A cognitive calculus, put simply, is a
logical system in which $\mathcal{L}$ includes intensional operators
(e.g.\ for such things as belief, desire, intention, emotional states,
communication, perception, and attention); $\mathcal{I}$ includes at
least one inference schema that involves such operators; and the
meaning of formulae, determined by some particular $\mathcal{S}$,
because they can in their expressive power far outstrip any standard,
off-the-shelf semantics (such as possible-worlds semantics), is
generally proof-theoretic in nature.

\subsection{The Learning \textit{Ex Nihilo} Loop}
\label{subsect:the_basic_loop}

Learning \textit{ex nihilo} happens when an agent loops through time,
as follows in broad strokes: Identify Interest/Goal $\Rightarrow$
Query $\Rightarrow$ Discover Argument/Proof to Answer
Query$\Rightarrow$ Learn $\Rightarrow$ Identify Interest/Goal, etc.
This cycle is at work in the parables with which we began.  We do not
have space to detail this persistent process.  In particular, the
management of the agent's interests (or goals) requires planning ---
but the emphasis in the present paper, for economy, is on reasoning.
Below we do discuss not only the AI technology that brings this loop
to life, but some simulation of the process in our earlier parables.


\section{The Automation of Learning \textit{Ex Nihilo}}
\label{sect:automation}

But how do we pass from the abstract logico-mathematics of learning
\textit{ex nihilo} to an artificial agent that can bring such a thing
to life?  The answer should be quite obvious, in general: We build an
AI that can find the arguments that undergird the knowledge obtained
by learning \textit{ex nihilo}.  In turn, this means that we need an
automated reasoner of sufficient power and reach that can pursue
epistemic interests, and a planner that can at least manage (e.g.\
prioritize) interests.  This brings us to the following progression,
in which we now briefly describe one such reasoner (ShadowReasoner),
and then give an illustrative simulation made possible by this AI
technology.

\subsection{Automated Reasoner:  ShadowReasoner}
\label{subsect:shadowprover}
A large amount of research and engineering has gone into building
first-order theorem provers in the last few decades.  ShadowReasoner
leverages this progress by splitting any logic into a first-order core
and the ``remainder,'' and then calls a first-order theorem prover
when needed.  Briefly, ShadowReasoner splits the inference schemata
for a given $\mathscr{L} \equiv \langle \mathcal{L}, \mathcal{I},
\mathcal{S} \rangle$ into two parts $\mathcal{I}_1$ and
$\mathcal{I}_{>1}$.  The first part $\mathcal{I}_1$ consists of
inference schemata that can be carried out by a first-order theorem
prover when the input expressions are \emph{shadowed} down into
first-order logic.  The second part consists of inference schemata
that cannot be reduced to first-order reasoning.  Given any problem in
a logic, ShadowReasoner alternates between trying out
$\mathcal{I}_{>1}$ and calling a first-order theorem prover to handle
$\mathcal{I}_1$.

The core algorithm for ShadowReasoner has a theoretical justification
based on the following theorem (which arises from the fact that
first-order logic can be used to simulate Turing machines
\cite{boolos_jeffrey_5thed}):

\begin{footnotesize} \begin{mdframed}[linecolor=white,
    frametitle=Theorem 1,frametitlebackgroundcolor=gray!25,
    backgroundcolor=gray!10]
  Given a collection of Turing-decidable inference schemata
  $\mathcal{I}$, for every inference $\Gamma\vdash_{\mathcal{I}}\phi$,
  there is a corresponding first-order inference $\Gamma'\vdash\phi'$.
\end{mdframed}
\end{footnotesize}

We illustrate how ShadowReasoner works in the context of a first-order
modal logic employed in \cite{nsg_sb_dde_ijcai}.  Please note though
there are some extant first-order modal-logic theorem provers, built
upon reductions to first-order theorem provers, they have some
deficiencies.  Such theorem provers achieve their reduction to
first-order logic via two methods.  In the first method, modal
operators are represented by first-order predicates.  This approach is
computationally fast but can quickly lead to well-known
inconsistencies, as demonstrated in
\cite{selmer_naveen_metaphil_web_intelligence}.  In the second method,
the entire proof theory is implemented in first-order logic, and the
reasoning is carried out \emph{within} first-order logic.  Here, the
first-order theorem prover simply functions as a programming system.
The second method, while accurate, can be excruciatingly slow.

ShadowReasoner uses the different approach alluded to above --- one in
which it alternates between calling a first-order theorem prover and
applying non-first-order inference schemata.  When we call the
first-order prover, all non-first-order expressions are converted into
propositional atoms (i.e., shadowing), to prevent substitution into
non-first-order contexts, as such substitutions lead to
inconsistencies \cite{selmer_naveen_metaphil_web_intelligence}.  This
approach achieves speed without sacrificing consistency.  The
algorithm is briefly described below.

First we define the syntactic operation of \textbf{atomizing} a
formula, denoted by $\mathsf{A}$.  Given any arbitrary formula $\phi$,
$\mathsf{A}_{[\phi]}$ is a unique atomic (propositional) symbol.
Next, we define the \textbf{level} of a formula: $\mathsf{level}:
\mathcal{L} \rightarrow \mathbb{N}$.

\begin{footnotesize}
  \begin{equation*}
    \begin{aligned}
      \mathsf{level}(\phi) = \left\{ 
        \begin{aligned}
          &0 ; \phi  \mbox{ is purely propositional
            formulae; e.g. } \mathit{Rainy} \\
          &1 ; \phi  \mbox{ has first-order
            predicates or quantifiers e.g. } \mathit{Sleepy}(\mathit{jack}) \\
          &2 ; \phi  \mbox{ has modal formulae e.g. } \knows(\mathit{a}, t, \mathit{Sleepy}(\mathit{jack}))
        \end{aligned}
      \right.
    \end{aligned}
  \end{equation*}
\end{footnotesize}

Given the above definition, we can define the operation of
\textbf{shadowing} a formula to a level.

\begin{footnotesize}
  \begin{mdframed}[linecolor=white, frametitle=Shadowing,
    frametitlebackgroundcolor=gray!25, backgroundcolor=gray!10,
    roundcorner=8pt] To shadow a formula $\chi$ to a level $l$,
    replace all sub-formulae $\chi'$ in $\chi$ such that
    $\mathsf{level}(\chi')>l$ with $\mathsf{A}_{[\chi']}$
    simultaneously.  We denote this by $\mathsf{S}[\phi,l]$.

    For a set $\Gamma$, the operation of shadowing all members in the
    set is simply denoted by $\mathsf{S}[\Gamma,l]$.
  \end{mdframed}
\end{footnotesize}

\noindent Assume we have access to a first-order prover
$\mathbf{P}_{F}$.  For a set of pure first-order formulae $\Gamma$ and
a first-order $\phi$, $\mathbf{P}_{F}(\Gamma, \phi)$ gives us a proof
for $\Gamma\vdash_{\mathcal{I}_1}\phi $ if such a first-order proof
exists; otherwise $\mathbf{P}_{F}$ returns \textsf{NO}.

\begin{footnotesize}
  \begin{algorithm}
\footnotesize
    \KwIn{Input Formulae $\Gamma$, Goal Formula $\phi$}
    \KwOut{A proof of $\Gamma\vdash_{\mathcal{I}}\phi$ if such a proof exists,
      otherwise \color{red}\textbf{fail}}
    initialization\;

    \While{goal not reached}{
      $ \mathit{answer}= \mathbf{P}_F(\mathsf{S}[\Gamma,1], \mathsf{S}[\phi,1])$\;

      \eIf{$\mathit{answer}  \not=\mathsf{NO}$}{
        \color{blue}{return} \color{black}$\mathit{answer}$ \;
      }{
        $\Gamma' \longleftarrow$ \textbf{expand} $\Gamma$ by using any
        applicable $\mathcal{I}_{>1}$\;
        \eIf{$\Gamma' = \Gamma$}{ \tcc{ The input cannot be expanded further}\ \color{blue}{return}
          \color{black} \color{red}\textbf{fail}}
        {set $\Gamma \longleftarrow \Gamma'$}
      }
    }
    \caption{ShadowReasoner Core Algorithm}
  \end{algorithm}

\end{footnotesize}




\subsection{Illustrative Simulation}
\label{subsect:simulation}

\begin{figure}
 \centering
\shadowbox{
  \includegraphics[width=\linewidth]{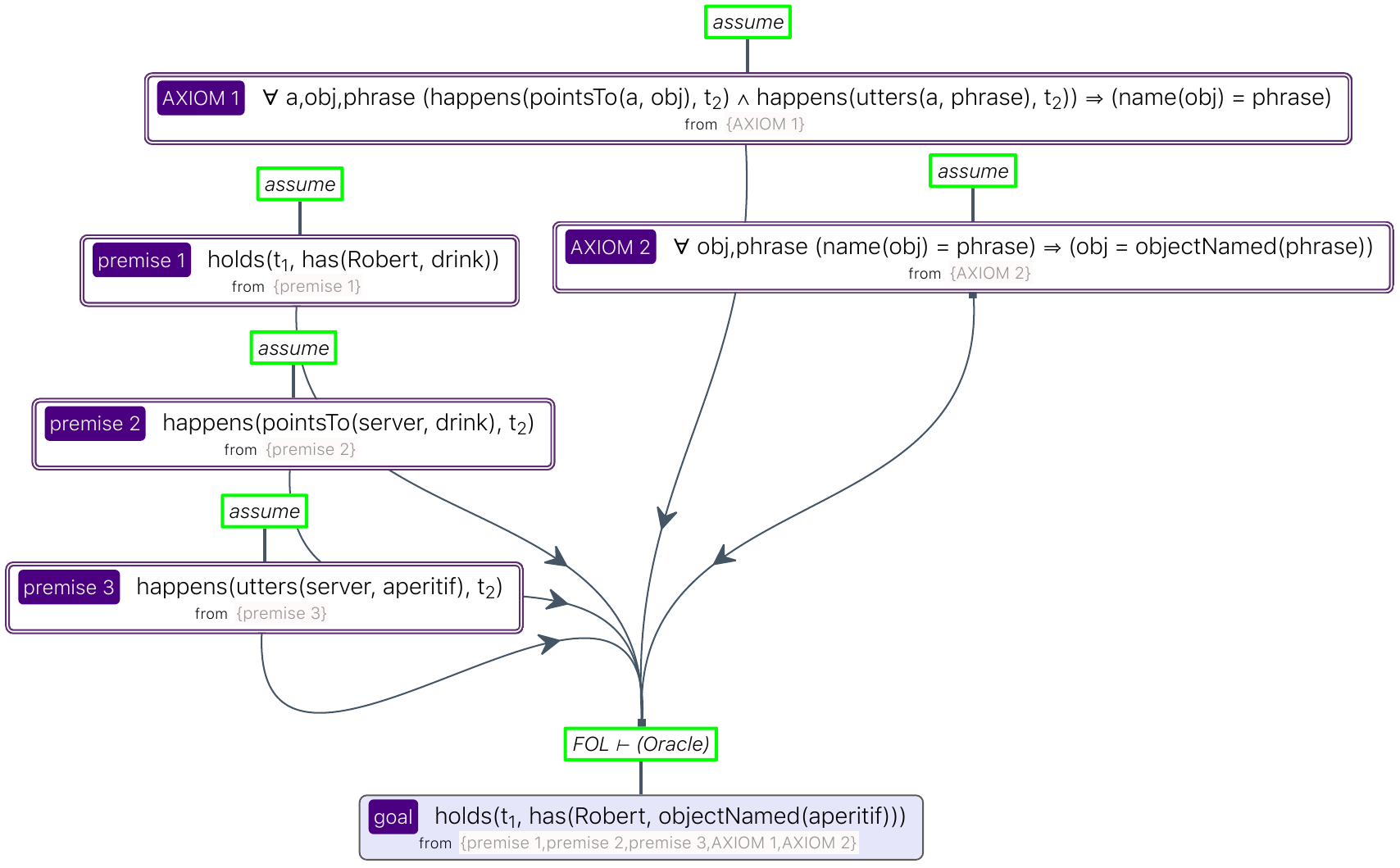}}
 \caption{Dinner Party Example Part 1}
 \label{fig:dpex1}
\end{figure}

\begin{figure}
 \centering
\shadowbox{
  \includegraphics[width=\linewidth]{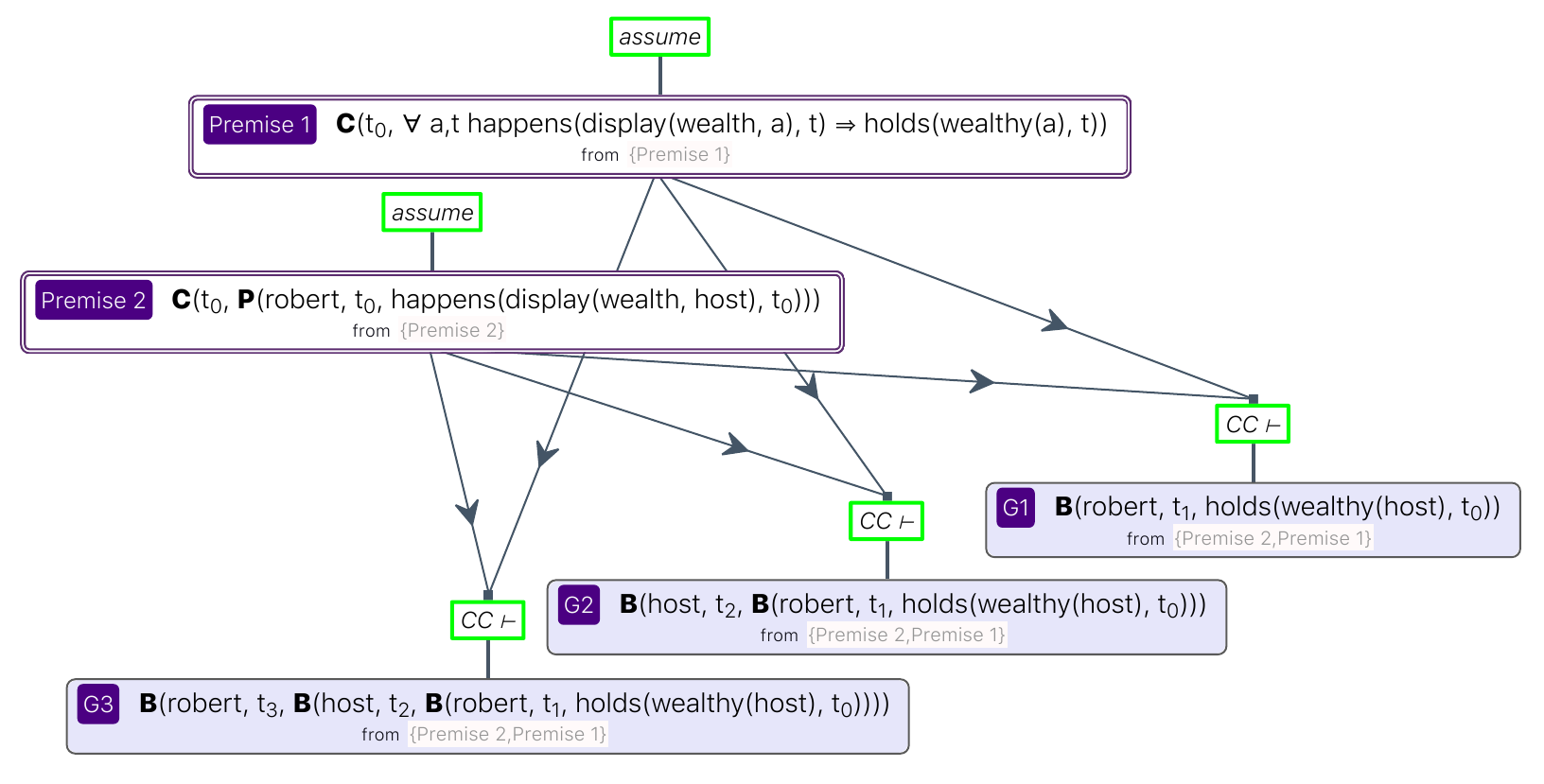}}
 \caption{Dinner Party Example Part 2 }
 \label{fig:dpex2}
\end{figure}

Figures~\ref{fig:dpex1} and~\ref{fig:dpex2} illustrate the
dinner-party parable simulated in the \textbf{deontic cognitive event
  calculus} (\DCEC) using ShadowReasoner within a graphical
argument-construction system; see \cite{Slate_at_CMNA08} for a
similar, but less intelligent, system. See the appendix for a
description of syntax and inference schemata of \DCEC.
Figure~\ref{fig:dpex1} simulates in pure first-order logic Robert's
learning that his drink is an aperitif.  Figure~\ref{fig:dpex2} is a
proof in a cognitive calculus, viz.\ the one described in
\cite{nsg_sb_dde_ijcai}, of Robert learning the following
propositions: \emph{``Robert believes that his host is wealthy''},
\emph{``The host believes Robert believes that his host is wealthy''},
and \emph{``Robert believes that his host believes Robert believes
  that his host is wealthy.''}\footnote{ With background information
  $t_i < t_j \mbox{ if } i<j$.}  The figures illustrate first-order
and cognitive-calculus reasoners (shown as $\mathsf{FOL} \vdash$ and
$\mathsf{CC} \vdash$, resp.)~being employed to derive these
statements.  Automated discovery of the proofs in~\ref{fig:dpex1} took
$\sim 0.1 (ms)$, and the proofs in~\ref{fig:dpex1} took
$\sim 0.9 (s)$. We briefly explain the figures now. The two figures
show assumptions that are fed to the reasoner and output formulae that
are proved by the reasoner, as denoted by boxes containing the
$\vdash$ symbols and the directions of the arrows. Each formula also
has a human readable label displayed with a purple background. The
boxes with shaded backgrounds are the outputs, the shading is not
necessary but makes it visually easier to see the outputs.  The text
below each formula shows what assumptions the formula is derived from
or dependent upon.

\section{++++ Objections, Rebuttals}
\label{sect:objections}

We now anticipate and rebut ++++ objections likely to come from
skeptics, including specifically those immersed in forms of learning
far removed from any notion of machine-verifiable proof or argument
enabled by inference schemata.

\subsection{Objection 1:  This isn't learning from \emph{nothing}!}
\label{subsect:from_nothing?}

The first objection is that `learning \textit{ex nihilo}' is a
misnomer.  The rationale for this complaint is simply the reporting of
an observation that should be clear to all: viz.\ that the learning in
question undeniably makes use of pre-existing stuff; hence we're not
dealing with learning
\emph{from nothing}.  In the parable of the dinner party, e.g., Robert
has brought his pre-existing command of elementary arithmetic to the
table; ditto for much other pre-known propositional content.  So how
then is it fair to speak of learning \textit{ex nihilo}?  It's fair
because obviously learning \textit{ex nihilo} trades on the
pre-existing concept of creation \textit{ex nihilo}, and that
millennia-old conception allows that the Almighty (by definition!)~was
around before the creation in question occurred.  And of course this
is just one part of pre-creation stuff in creation
\textit{ex nihilo}: God presumably needed to have a mental concept of
a planet to create a planet.  We generally suspect that
learning \textit{ex nihilo} begins to kick into ``high gear'' in the
human sphere when children are sophisticated enough to exploit their
prior knowledge of content that requires, for its underlying
representation, $\mathcal{L}_1$ and basic modal logic.  From the
standpoint of logicist cognitive science, rather than AI, this means
that learning \textit{ex nihilo} would be aligned with the views of
Piaget and colleagues [e.g.\ \cite{inhelder.piaget}], Stenning and
colleagues [e.g.\ \cite{stenning_reasoning_cogsci}], Bringsjord and
colleagues [e.g.\ \cite{logical.minds,rascals_in_sun}], and Rips and
colleagues [e.g.\ \cite{rips.book,rips_knights_knaves_1989}].  The
goal of full formalization and implementation of learning \textit{ex
nihilo} would likely be regarded by these cognitive scientists as a
welcome one.

\subsection{Objection 2:  Isn't this just reasoning?}
\label{subsect:just_reasoning?}

The second objection we anticipate is that learning \textit{ex nihilo}
isn't learning at all; it's just a form of reasoning.  In reply, any
process, however much it relies on reasoning, that does enable an
agent running that process to acquire genuine knowledge (and
our \texttt{j=tb} definition of knowledge, note, is a very demanding
one) would seem to be quite sensibly classified as a learning process.
In fact, probably it strikes many as odd to say that one has a form of
machine learning that does \emph{not} result in the acquisition of any
knowledge.


\subsection{Objection 3: What about inductive logic programming?}
\label{subsect:what_about_ILP}

The objection here can be encapsulated thus: ``What about inductive
logic programming (ILP)?  Surely this established family of techniques
both uses formal logic, and results in new knowledge.''

ILP is along the general lines of learning \textit{ex nihilo}, agreed;
but ILP is acutely humble and highly problematic when measured against
LEN, for reasons we give momentarily.  Before giving these reasons,
without generaity we fix a general framework $\mathcal{F}$
\citeasnoun{mooney_integrating_abduction_induction} to fix abduction
and induction in supposedly representative logicist fashion:

  \begin{small}
    \begin{quote}
      \begin{description}

        \item[Given:] Background knowledge, $B$, and observations
          (data), $O$, both represented as sets of formulae in
          first-order predicate calculus, where $O$ is restricted to
          ground formulae.

        \item[Find:] An hypothesis $H$ (also a set of logical
          formulae) such that $B \cup H \not\vdash \bot \mbox{ and } B
          \cup H \vdash O$.

      \end{description}
    \end{quote}
  \end{small}

\noindent From $\mathcal{F}$ one can derive both induction and
abduction, according to Mooney: For the latter, $H$ is restricted to
atomic formulae or simple formulae of the general form $\exists x
\phi$, and $B$ is --- as Mooney says --- ``significantly larger'' than
$H$.  For induction, Mooney says that $H$ is to consist of universally
quantified Horn clauses, and $B$ is ``relatively small'' and may even
be the null set.

From an explain-the-extant-literature point of view, $\mathcal{F}$ is
at least somewhat attractive.  For as
\citeasnoun{mooney_integrating_abduction_induction} points out, this
framework captures what many logic-oriented learning researchers in AI
have taken induction and abduction to be; this includes, specifically,
ILP researchers \citeaffixed{ilp_book,ilp_lavrac}{e.g.}.
Unfortunately, despite its explanatory virtue, $\mathcal{F}$, when
measured against human cognition of the sort involved in learning
\textit{ex nihilo}, is embarrassingly inadequate.  As we have said,
this can be shown effortlessly via many reasons.  We quickly mention
just four here.

  \subsubsection{Reason 1: $\mathcal{F}$ Runs Afoul of Non-Deductive Reasoning}
  \label{subsubsect:reason1}

  Why should the combination of background knowledge and a candidate
  hypothesis need to \emph{deductively entail} some observation, as
  $\mathcal{F}$ says (via its use of $\vdash$)?  Suppose that as Smith
  sits in his home office looking through a window he perceives
  ($\omega$) that a uniformed man with a \textsc{FedEx} cap is
  approaching Smith's house, a small white box in hand.  Smith has no
  doubt learned that ($\delta$) a delivery is about to be attempted,
  but does it really need to be the case that, where $B$ is background
  knowledge, $B \cup \{\omega\}$ can be used to \emph{prove} $\delta$?
  No, it doesn't.  Maybe it's Halloween, Smith forgot that it is, and
  the person approaching is in costume and playacting.  Maybe the
  approaching man is a criminal in disguise, merely casing Smith's
  domicile.  And so on.  All that is needed is a fairly strong
  \emph{argument} in support of $\delta$.  And that argument can make
  use of inferences that are deductively invalid, but valid as
  reasoning that is analogical, adbuctive, inductive,
  etc.\footnote{Indeed, these inferences can even be \emph{formally}
    valid in the inductive logics that will undergird ALML; see
    below.}

  \subsubsection{Reason 2: $\mathcal{F}$ Leaves Aside Other Non-Deductive Reasoning}
  \label{subsubsect:reason1}

  This reason was revealed in the previous sentence: that sentence
  refers to not just to abduction and induction, but also to reasoning
  that is \emph{analogical} in nature, and such reasoning isn't
  included in ILP.  In fact, learning via analogical reasoning is
  often left aside in coverage of logicist learning of \emph{any}
  textbook variety.\footnote{E.g.\ learning by analogy isn't included
    in AI's definitive, otherwise encyclopedic textbook:
    \cite{aima.third.ed}.}  And as we have seen above, even Pollock
  mentions only abduction and induction; he leaves analogical
  reasoning aside.  But if in an earlier case Smith had encountered
  not a \textsc{FedEx} man, but rather a USPS mailman making a
  delivery to his house, he may well have believed that the man with
  the \textsc{FedEx} cap was analogous, and hence was making a
  delivery.

  \subsubsection{Reason 3: $\mathcal{F}$ Runs Afoul of Robust Expressivity}
  \label{subsubsect:reason3}

  A quick but decisive third reason Mooney's $\mathcal{F}$ explodes in
  the face of real human cognition is that any expressivity
  restriction on $O$ and/or $H$ is illogical, and certainly any
  specific restriction that $O$ be restricted to ground formulae
  and/or that $H$ be confined to Horn-clause logic (or even for that
  matter full FOL) is \emph{patently} illogical.  This can be seen in
  middle-school classes that cover arithmetic, where even very young
  students cook up and affirm hypotheses that are expressed using
  infinitary constructions beyond even full FOL.  For instance:
  Student Johnny is reminded of the definition of a prime number, and
  then shown that 4 is equal to 3 plus 1, that 6 is equal to $5 + 1$,
  that $8 = 5 + 3$, etc.  Johnny is asked to consider whether the next
  two even numbers continue the pattern.  He observes that $10 = 7 +
  3$ and that $12 = 9 + 3$, and is now inclined to hypothesize ($H'$)
  that every even integer greater than 2 is the sum of two primes.  A
  natural form for $H'$, where $e_1$ is 2 and the even numbers from
  there continue $e_2, e_3, \ldots$ is simply a list $L$ like:
  $$e_1 = p_1 + p'_1$$
  $$e_2 = p_2 + p'_2$$
  $$\vdots$$
  Yet $L$ cannot be expressed in finitary formulae,\footnote{It is
    naturally represented by an infinite conjunction, which the logic
    $\mathcal{L}_{\omega_1 \omega}$ allows.} and even if one squeezed
  $L$ into finitary FOL, that would be done by employing the same
  trick as is used for instance in axiomatic set theory, where FOL is
  augmented with schematic formula that denote an infinite number of
  instantiations thereof.\footnote{E.g., witness the Axiom
    \emph{Schema} of Separation as a replacement for Frege's doomed
    Axiom V, shot down violently and decisively by the famous
    Russell's Paradox; see \cite{suppes_axiomatic_set_theory} for
    wonderfully elegant coverage.} Regardless, even if by some miracle
  $H'$ could be expressed in some finitary extensional logic at or
  beneath FOL, the classroom in question wouldn't exactly operate well
  without the teacher's believing Johnny's believing that $H'$, and
  certainly nothing like this fact is expressible in extensional logic
  (let alone Horn-clause logic!).

  \subsubsection{Reason 4: $\mathcal{F}$ Ignores Thinking}
  \label{subsubsect:reason4}

  The dictum that truth is stranger than fiction, alas, is frequently
  confirmed by the oddities of contemporary AI research.  The fourth
  reason $\mathcal{F}$ is inadequate is a confirming example.  For
  while the framework $\mathcal{F}$ projects an air of being about
  thinking, in actuality it leaves thinking aside.  Indeed there's a
  rather common fallacy at work in $\mathcal{F}$, and its promotion:
  the fallacy of composition.  For while $\mathcal{F}$ does include
  some forms of reasoning, these forms (and even, for that matter,
  \emph{all} forms of reasoning put together) don't comprise thinking;
  thinking includes the reasoning called out in $\mathcal{F}$ as
  merely a \emph{proper part}.  To be more specific, in real and
  powerful thinking, an hypothesis $H$ can be wisely discarded when
  there is evidence against it.  (The rejection of the aether drag
  hypothesis is a firm case in point.)

  The fourth flaw infecting $\mathcal{F}$ can be easily unpacked: Real
  learning is intertwined with the full gamut of human-level thinking:
  with planning, decision-making, and communication.  If there are no
  observations to learn from, an agent can plan to get more
  observations.  An agent can decide when to learn and what to learn.
  Precious little substantive learning takes place without
  communication between teacher and learner, including written
  communication.  And finally, at the highest end of the spectrum of
  powerful learning, learners formalize learning itself, and learn
  more by doing so.

\subsection{Objection 4: Does ShadowReasoner  really conform to the K=jtb thesis?}

\begin{quote}
``It is not clear for me how and why the approach produces \textbf{true}
beliefs. Indeed, it seems to me that the reasoning that is performed
by ShadowReasoner is internal to it and that propositions that are
'produced' are not necessarily true. In other words, why does
ShadowReasoner conform to the K=jtb thesis?''
\end{quote}

There is a tradition in agent-based modeling in which there is
a difference between a system S (e.g. ShadowReasoner) and the system S
modeling other agents. So truth is what the system knows, and so that
knowledge that P on the part of an agent is inferred from that
agent believing P, knowing that some proof/argument for P exists,
and the system's knowing that P.

\subsection{Objection 5: Isn't this just deduction}
\label{subsect: obj_analogical_schema.tex}

Cognitive calculi have been used to capture and model non-deductive inference systems. 
See \cite{cyber_new_disanalogy_phil_tech} for one such example.


\subsection{Objection 6: What about Lewis' ``Effusive Knowledge''?}
\label{subsect:lewis_effusive}

``You claim to have a solution to the Gettier problem, one seemingly
based on introducing different levels of knowledge of $p$ (why five?),
based on the different levels of justification one may have for $p$.
This idea echoes Lewis’ \citeyear{lewis_elusive_knowledge} epistemic
contextualism,
but does it really solve the Gettier problem?  Anbd I don't see any
justification for why, say, level-5 \texttt{jtb} is to be equated with
full knowledge?''

Actually, Lewis' ``Effusive Knowledge'' paper espouses a conception
that is the \emph{opposite} of what underlies learning \textit{ex
  nihilo}.  Lewis holds that there is no knowledge in the Gettier
scenarios, because his definition for knowledge (which marks a
rejection of \texttt{k=jtb}) isn't satisfied in these cases.  Learning
\textit{ex nihilo} entails that there \emph{is} knowledge in these
scenarios, but \emph{reduced} knowledge.  If the degree of belief is
$k$, then knowledge partaking of this belief is of degree $k$ as well.
In Gettier's original scenarios, knowledge is at the level of 2 (for
reasons too far afield to articulate under current space constraints).
It seems never to have occurred to Lewis that if belief comes in
degrees, then knowledge (which surely includes
belief\footnote{Actually, Lewis asserts that there can be knowledge
  without belief, because a timid student can know the answer, but has
  ``no confidence that he has it right, and so does not believe what
  he knows'' (p.\ 556).  In our formal framework, the situation is
  that the student believes, at a low level (1, say) that he knows (at
  some level $k \ge 1$) the answer, and as a matter of fact he
  \emph{does} know the answer at level 1.  This scenario is provably
  consistent in our relevant cognitive calculi.  Not only that, but as
  far as we can tell, since in point of fact timidity of this type
  often prevents successful performance, our formal diagnosis has
  ``real-worl'' plausibility.})  must itself come in degrees.  In the
context of formal intensional logic (e.g.\ formal epistemic logic),
Lewis position/paper is from our formalist point of view weak, because
it hinges on the repeatedly asserted-without-argument claim that in
the case of a single-ticket lottery of size $m$, even when the number
of tickets is aribrarily large (eg $m$ = 1 quadrillion), one cannot
know that a given ticket $t_k$ ($1 \le k \le 1Q$) will not win.  From
the standpoint of learning \textit{ex nihilo}, this is bizarre,
because surely one knows that in the next moment one's computer will
not spontaneously combust, because such an event is preposterously
unlikely.  (And looking back in time, surely one knows that the
computer sitting here now was there 10 seconds ago.)  Moreover, Lewis
rejects the very concept of justification as part and parcel of
knowledge; learning \textit{ex nihilo} by contrast is an AI-driven
conception based on automated reasoning (automated deduction and
automated analogical, abductive, enumerative inductive etc reasoning).

As to levels of belief, in the case of an 11-valuded inductive logic
we use to undergird learning \textit{ex nihilo}, 5 = \textit{certain},
4 = \textit{overwhelmingly likely}, 3 = \textit{beyond reasonable
  doubt}, 2 = \textit{likely}, 1 = \textit{more likely than not}.  0
is \textit{counterbalanced}, and then we have the symmetrical negative
values.  A belief at level 5 corresponds to knowing things that Lewis
(and everyone else) agree that we know: e.g.\ knowing that 2+2=4.

\subsection{Objection 7: What about QMLTP and isn't shadowing similar
  to existing schemes?}
\label{subsect:obj_shadowing}

Note: we deal with /indexed/ modal operators (agent, time and formulae for dyadic obligations).  We can have aribtary expressions for these indices. Our goals and systems are much different from QMLTP \cite{Raths:2012:QPL:2352896.2352933}.  Schemes such as the Standard Translation for modal logics are semantic in nature and apply only for propositional logics.

\subsection{Objection 8: Examples are too simple and the discussion is
too philosophical}
\label{subsect:deeper_example}

Our desire was to introduce, mostly philosophically, and to a degree
 formally, LEN.  We also point out connections to famous fictional
 detection, and cognitive psychology (learning by posing questions to
 oneself, which is seminal work by Chi et al).  A purely formal
 treatment can we believe follow quite naturally after this
 introduction, and fit the formal elements we already give.  Such a
 treatment includes further specification of cognitive calculi where
 the inference schemata $\mathcal{I}$ run the gamut of inductive logic;
 mechanisms for question-generation; mechanisms for argument-discovery
 over these inference schemata by automated-reasoning technques that
 cover all major forms of reasoning; and an argument-checker to
 validate discovered arguments.  ShadowReasoner is the current
 implemented system used for argument- and proof-discovery, and
 checking.  The full logico-mathematics of LEN, and corresponding
 implementation, is something we now feel we should give more of
 (thank you), but we are concerned that jumping further in that
 direction, without seeking to communicate to those in ML who don't do
 formal logic, might retard cross-paradigm collaboration on learning.
  In our experience, many stat ML folks don't know e.g. what a logical
 system is even in the standard extensional sense we start with
 (Lindstrom), and don't know how it might connect to everyday human
 activity.  Since in the general case argument- or proof-discovery in
 a cognitive calculus is far above what's Turing-computable, our
 strategy was to try to connect to "everyday" parables first.  As far
 as we can tell, the functions learned by systems doing ML are
 Turing-computable; we feel a special need to begin with such parables
 in order to show the learning by reasoning in question isn't rare,
 and apologize if we have miscalculated here.

 LEN comes with a novel solution to the Gettier problem that has
 plagued (if not eviscertated) larges parts of epistemology in
 philosophy for decades --- and at the same time insists upon a form
 of ML that produces propositional knowledge.  This is a form of
 knowledge that Gettier cases imperil.

 Perhaps more importantly, we are concerned that ML folks of the
 statistical/connectionist variety will not engage more
 straightforwardly technical content.  In our experience, while you
 know and deal deeply with higher-order logics, stat/conn ML folks
 don't even know why zero-order logic might be relevant to ML, or AI
 in general.  We may be miscalculating, but we are seeking to provide
 informal, philosophical toeholds to bring somme of these people into
 collaboration.

\subsection{Objection n:  Frivolity?}
\label{subsect:frivolity}

Finally, some will doubtless declare that learning \textit{ex nihilo}
is frivolous.  What good is it, really, to sit at a dinner table and
learn useless tid-bits?  This attitude is most illogical.  The reason
is that, unlike what is currently called ``learning,'' only persons at
least at the level of humans can learn \textit{ex nihilo}, and this
gives such creatures power, for the greatest of what human persons
have learned (and, we wager, \emph{will} learn) comes via learning
\textit{ex nihilo}.  In support of this we could cite countless
confirmatory cases in the past, but rest content to but point out that
armchair learning \textit{ex nihilo} regarding simultaneity (Einstein)
and infinitesimals (Leibniz) was rather productive.\footnote{And for
those readers with a literary bent, it should also be pointed out that
the great minds of detection, not only the aforementioned Sherlock
Holmes but e.g.\ Poe's seminal Le Chevalier C.\ Auguste Dupin, achieve
success primarily because of their ability to learn \textit{ex
nihilo}.}

\section{Conclusion and Next Steps}
\label{sect:conclusion}

We have provided an introduction, philosophical in nature, to the new
concept of learning \textit{ex nihilo}, and have included enough
information re.\ its formal foundation to allow those conversant with
logicist AI to better understand this type of learning.  In addition,
we have explained that learning \textit{ex nihilo} can be automated
via sufficiently powerful automated-reasoning technology.  Of course,
this is a very brief paper.  Accordingly, next steps include
dissemination of further details, obviously.  But more importantly,
what is the relationship between learning \textit{ex nihilo} and types
of machine learning that are based on artificial neural networks,
Bayesian reasoning, reinforcement learning, and so on?  These are
other currently popular types of learning are certainly \emph{not}
logicist, and hence nothing like a logical system, let alone a
cognitive calculus, are present.  In fact, it's far from clear that
it's even possible to construct the needed machinery for learning
\textit{ex nihilo} out of the ingredients that go into making these
non-logicist forms of learning.

\begin{small}
  \paragraph{Acknowledgements} We are deeply grateful for current
  support from the U.S.\ Office of Naval Research to invent,
  formalize, and implement new forms of learning based on automated
  reasoning.  Prior support from DARPA of logicist learning has also
  proved to be helpful, and for this too the first author expresses
  thanks.  Four anonymous reviewers provided insightful feedback, for
  which we are deeply grateful.
\end{small}

\clearpage

\bibliographystyle{agsm}
\balance
    \bibliography{main72edited,naveen}

\appendix
\section{Deontic Cognitive Event Calculus}

\DCEC\ is a quantified multi-modal sorted calculus and a cognitive
calculus. A sorted system can be regarded analogous to a typed
single-inheritance programming language. We show below some of the
important sorts used in \DCEC. Among these, the \type{Agent},
\type{Action}, and \type{ActionType} sorts are not
native to the event calculus.\\



\begin{scriptsize}
\begin{tabular}{lp{5.8cm}}  
\toprule
Sort    & Description \\
\midrule
\type{Agent} & Human and non-human actors.  \\

\type{Time} &  The \type{Time} type stands for
time in the domain.  E.g.\ simple, such as $t_i$, or complex, such as
$birthday(son(jack))$. \\

 \type{Event} & Used for events in the domain. \\
 \type{ActionType} & Action types are abstract actions.  They are
  instantiated at particular times by actors.  Example: eating.\\
 \type{Action} & A subtype of \type{Event} for events that occur
  as actions by agents. \\
 \type{Fluent} & Used for representing states of the world in the
  event calculus. \\
\bottomrule
\end{tabular}
\end{scriptsize} \\

 Note: actions are events that are carried out by an agent. For any
action type $\alpha$ and agent $a$, the event corresponding to $a$
carrying out $\alpha$ is given by $\action(a, \alpha)$. For instance
if $\alpha$ is \textit{``running''} and $a$ is \textit{``Jack'' },
$\action(a, \alpha)$ denotes \textit{``Jack is running''}.

\subsection{Syntax}
\label{subsect:syntax}

The syntax has two components: a first-order
core and a modal system that builds upon this first-order core.  The
figures below show the syntax and inference schemata of \DCEC.  The
syntax is quantified modal logic. The first-order core of \DCEC\ is
the \emph{event calculus} \cite{mueller_commonsense_reasoning}.
Commonly used function and relation symbols of the event calculus are
included.  Other calculi (e.g.\ the \emph{situation calculus}) for
modeling commonsense and physical reasoning can be easily switched out
in-place of the event calculus.



The modal operators present in the calculus include the standard
operators for knowledge $\knows$, belief $\believes$, desire
$\desires$, intention $\intends$, etc.  The general format of an
intensional operator is $\knows\left(a, t, \phi\right)$, which says
that agent $a$ knows at time $t$ the proposition $\phi$.  Here $\phi$
can in turn be any arbitrary formula. Also,
note the following modal operators: $\mathbf{P}$ for perceiving a
state, 
$\mathbf{C}$ for common knowledge, $\mathbf{S}$ for agent-to-agent
communication and public announcements, $\mathbf{B}$ for belief,
$\mathbf{D}$ for desire, $\mathbf{I}$ for intention, and finally and
crucially, a dyadic deontic operator $\mathbf{O}$ that states when an
action is obligatory or forbidden for agents. It should be noted that
\DCEC\ is one specimen in a \emph{family} of easily extensible
cognitive calculi.
 
The calculus also includes a dyadic (arity = 2) deontic operator
$\ought$. It is well known that the unary ought in standard deontic
logic lead to contradictions.  Our dyadic version of the operator
blocks the standard list of such contradictions, and
beyond.\footnote{A overview of this list is given lucidly in
  \cite{sep_deontic_logic}.}

 \begin{scriptsize}
\begin{mdframed}[linecolor=white, nobreak, frametitle=Syntax,
  frametitlebackgroundcolor=gray!15, backgroundcolor=gray!05,
  roundcorner=8pt]
 \begin{equation*}
 \begin{aligned}
    \mathit{S} &::= 
    \begin{aligned}
      & \Agent \sep \ActionType \sep \Action \sqsubseteq
      \Event \sep \Moment  \sep \Fluent \\
    \end{aligned} 
    \\ 
    \mathit{f} &::= \left\{
    \begin{aligned}
      & \action: \Agent \times \ActionType \rightarrow \Action \\
      &  \initially: \Fluent \rightarrow \Boolean\\
      &  \holds: \Fluent \times \Moment \rightarrow \Boolean \\
      & \happens: \Event \times \Moment \rightarrow \Boolean \\
      & \clipped: \Moment \times \Fluent \times \Moment \rightarrow \Boolean \\
      & \initiates: \Event \times \Fluent \times \Moment \rightarrow \Boolean\\
      & \terminates: \Event \times \Fluent \times \Moment \rightarrow \Boolean \\
      & \prior: \Moment \times \Moment \rightarrow \Boolean\\
    \end{aligned}\right.\\
        \mathit{t} &::=
    \begin{aligned}
      \mathit{x : S} \sep \mathit{c : S} \sep f(t_1,\ldots,t_n)
    \end{aligned}
    \\ 
    \mathit{\phi}&::= \left\{ 
    \begin{aligned}
     & q:\Boolean \sep  \neg \phi \sep \phi \land \psi \sep \phi \lor
     \psi \sep \forall x: \phi(x) \sep \\\
 &\perceives (a,t,\phi)  \sep \knows(a,t,\phi) \sep     \\ 
& \common(t,\phi) \sep
 \says(a,b,t,\phi) 
     \sep \says(a,t,\phi) \sep  \believes(a,t,\phi) \\
& \desires(a,t,\phi)  \sep \intends(a,t,\phi) \\ & \ought(a,t,\phi,(\lnot)\happens(action(a^\ast,\alpha),t'))
      \end{aligned}\right.
  \end{aligned}
\end{equation*}
\end{mdframed}
\end{scriptsize}

\subsection{Inference Schemata}

The figure below shows the inference schemata for
\DCEC.  $R_\mathbf{K}$ and $R_\mathbf{B}$ are inference schemata that
let us model idealized agents that have their knowledge and belief
closed under the \DCEC\ proof theory.  While normal humans are not
dedcutively closed, this lets us model more closely how deliberate
agents such as organizations and more strategic actors reason. (Some
dialects of  cognitive calculi restrict the number of iterations on intensional
operators.)  $R_1$ and $R_2$ state respectively that it is common
knowledge that perception leads to knowledge, and that it is common
knowledge that knowledge leads to belief.  $R_3$ lets us expand out
common knowledge as unbounded iterated knowledge.  $R_4$ states that
knowledge of a proposition implies that the proposition holds.  $R_5$
to $R_{10}$ provide for a more restricted form of reasoning for
propositions that are common knowledge, unlike propositions that are
known or believed.  $R_{12}$ states that if an agent $s$ communicates
a proposition $\phi$ to $h$, then $h$ believes that $s$ believes
$\phi$.  $R_{14}$ dictates how obligations get translated into
intentions.

\begin{scriptsize}
\begin{mdframed}[linecolor=white, nobreak, frametitle=Inference Schemata, frametitlebackgroundcolor=gray!15, backgroundcolor=gray!05, roundcorner=8pt]
\begin{equation*}
\begin{aligned}
  &\infer[{[R_{\knows}]}]{\knows(a,t_2,\phi)}{\knows(a,t_1,\Gamma), \ 
    \ \Gamma\vdash\phi, \ \ t_1 \leq t_2} \hspace{10pt} \infer[{[R_{\believes}]}]{\believes(a,t_2,\phi)}{\believes(a,t_1,\Gamma), \ 
    \ \Gamma\vdash\phi, \ \ t_1 \leq t_2} \\
 &\infer[{[R_1]}]{\common(t,\perceives(a,t,\phi) \lif\knows(a,t,\phi))}{}\hspace{6pt}
  \infer[{[R_2]}]{\common(t,\knows(a,t,\phi)
    \lif\believes(a,t,\phi))}{}\\
  &\infer[{[R_3]}]{\knows(a_1, t_1, \ldots
    \knows(a_n,t_n,\phi)\ldots)}{\common(t,\phi) \ t\leq t_1 \ldots t\leq
    t_n}\hspace{10pt}
  \infer[{[R_4]}]{\phi}{\knows(a,t,\phi)}\\
  & \infer[{[R_5]}]{\common(t,\knows(a,t_1,\phi_1\lif\phi_2))
    \lif \knows(a,t_2,\phi_1) \lif \knows(a,t_3,\phi_2)}{}\\
& \infer[{[R_6]}]{\common(t,\believes(a,t_1,\phi_1\lif\phi_2))
    \lif \believes(a,t_2,\phi_1) \lif \believes(a,t_3,\phi_2)}{}\\
& \infer[{[R_7]}]{\common(t,\common(t_1,\phi_1\lif\phi_2))
    \lif \common(t_2,\phi_1) \lif \common(t_3,\phi_2)}{} \\
& \infer[{[R_8]}]{\common(t, \forall x. \  \phi \lif \phi[x\mapsto
  t])}{} \hspace{18pt}
  \infer[{[R_9]}]{\common(t,\phi_1 \liff \phi_2 \lif \neg
    \phi_2 \lif \neg \phi_1)}{}\\
& \infer[{[R_{10}]}] {\common(t,[\phi_1\land\ldots\land\phi_n\lif\phi]
  \lif [\phi_1\lif\ldots\lif\phi_n\lif\psi])}{}\\
&\infer[{[R_{12}]}]{\believes(h,t,\believes(s,t,\phi))}{\says(s,h,t,\phi)}
\hspace{18pt}\infer[{[R_{13}]}]{\perceives(a,t,\happens(\action(a^\ast,\alpha),t))}{\intends(a,t,\happens(\action(a^\ast,\alpha),t'))}\\
&\infer[{[R_{14}]}]{\knows(a,t,\intends(a,t,\chi))}{\begin{aligned}\ \ \ \ \believes(a,t,\phi)
 & \ \ \
 \believes(a,t,\ought(a,t,\phi, \chi)) \ \ \ \ought(a,t,\phi,
 \chi)\end{aligned}}
\end{aligned}
\end{equation*}
\end{mdframed}
\end{scriptsize}


\end{document}